\documentclass[runningheads]{llncs}
\usepackage{main}
\usepackage{mainabbrv}

\usepackage{graphicx}
\usepackage{booktabs}
\usepackage{multirow}
\usepackage{multicol}
\usepackage{tabularx}
\usepackage{pifont}
\usepackage{multirow}
\usepackage{subcaption}

\usepackage[marginal]{footmisc}

\usepackage[accsupp]{axessibility}

\usepackage[pagebackref,breaklinks,colorlinks,citecolor=eccvblue]{hyperref}

\begin{document}

\title{OneVOS: Unifying Video Object Segmentation with All-in-One Transformer Framework}

\titlerunning{ }

\author{Wanyun Li \inst{1,2,*} \and
Pinxue Guo\inst{3,*} \and Xinyu Zhou \inst{1,2} \and
Lingyi Hong \inst{1,2} \and Yangji He\inst{1,2} \and Xiangyu Zheng \inst{1,2} \and Wei Zhang \inst{1,2,\dag} \and Wenqiang Zhang  \inst{1,2,3,\dag}  }

\authorrunning{ }
\institute{School of Computer Science, Fudan University \and
Shanghai Key Laboratory of Intelligent Information Processing \and 
Academy for Engineering and Technology, Fudan University} 

\maketitle  

\begin{abstract}
Contemporary Video Object Segmentation (VOS) approaches typically consist stages of feature extraction, matching, memory management, and multiple objects aggregation. Recent advanced models either employ a discrete modeling for these components in a sequential manner, or optimize a combined pipeline through substructure aggregation. However, these existing explicit staged approaches prevent the VOS framework from being optimized as a unified whole, leading to the limited capacity and suboptimal performance in tackling complex videos. In this paper, we propose OneVOS, a novel framework that unifies the core components of VOS with All-in-One Transformer. Specifically, to unify all aforementioned modules into a vision transformer, we model all the features of frames, masks and memory for multiple objects as transformer tokens, and integrally accomplish feature extraction, matching and memory management of multiple objects through the flexible attention mechanism. Furthermore, a Unidirectional Hybrid Attention is proposed through a double decoupling of the original attention operation, to rectify semantic errors and ambiguities of stored tokens in OneVOS framework. Finally, to alleviate the storage burden and expedite inference, we propose the Dynamic Token Selector, which unveils the working mechanism of OneVOS and naturally leads to a more efficient version of OneVOS. Extensive experiments demonstrate the superiority of OneVOS, achieving state-of-the-art performance across 7 datasets, particularly excelling in complex LVOS and MOSE datasets with 70.1\% and 66.4\% $J \& F$ scores, surpassing previous state-of-the-art methods by 4.2\% and 7.0\%, respectively. And our code will be available for reproducibility and further research.
\footnote{*Equal Contribution \\ \dag Corresponding Author} 
\end{abstract}

\section{Introduction}
\label{sec:intro}

\begin{figure*}[t]
  \centering
   \includegraphics[width=0.9\linewidth]{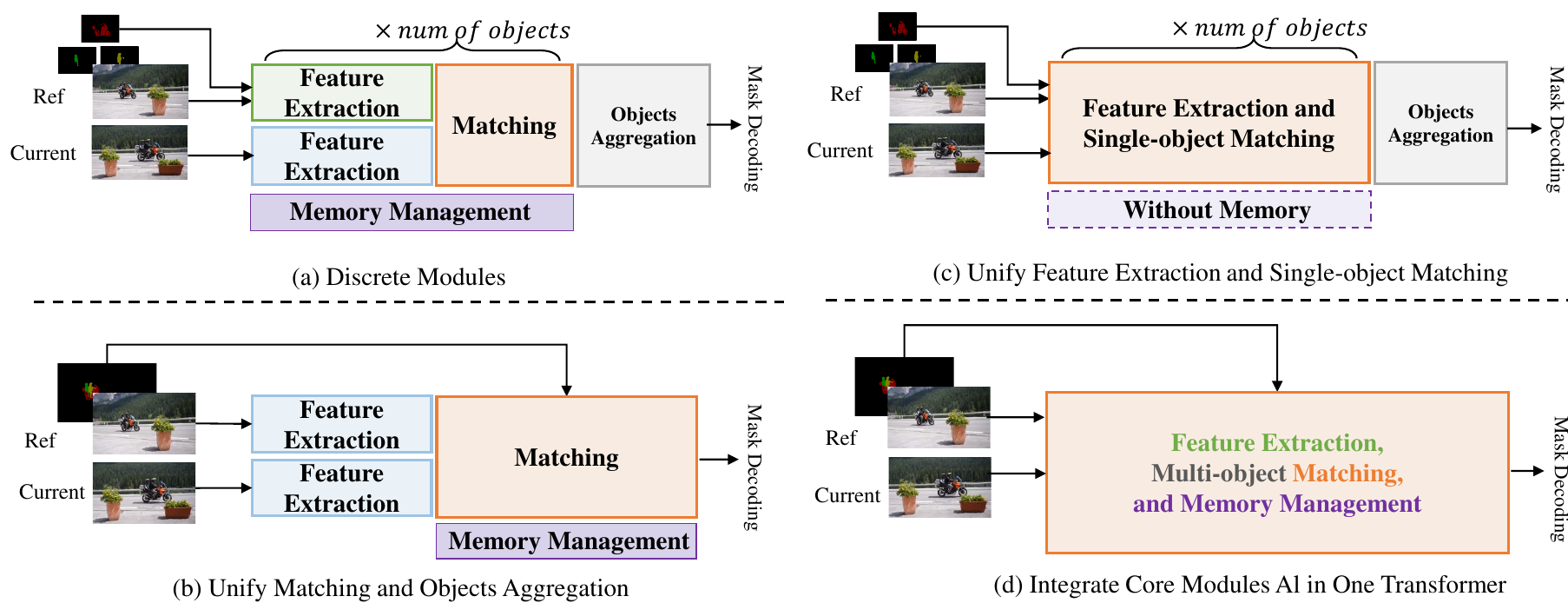}
    \caption{Four different modeling for semi-supervised video object segmentation pipelines. (a) Traditional discrete modeling of VOS. (b)  Unified multiple object modeling in one reference. (c) Unified single-object feature extraction and matching. (d) OneVOS: Integrates all core modules with All-in-One Transformer.}
   \label{fig:firstimg}
\end{figure*}

Semi-supervised Video Object Segmentation (VOS) \cite{caelles2017one,oh2018fast,oh2019video,hu2018videomatch}  is a fundamental task in video analysis, aiming to consistently segment objects throughout a video based on initial masks. Advancements in VOS primarily focus on the full utilization of available information. Among them, Memory-based methods\cite{oh2019video,lin2022swem,cheng2021rethinking,xmem} have emerged as the mainstream due to the effectiveness in storing and leveraging abundant spatio-temporal features. Its foundational paradigm, as depicted in Fig. 1(a) \cite{seong2021hierarchical,duke2021sstvos,lin2022swem,wang2021swiftnet}, encompasses four split key stages including feature extraction, matching, memory management, and multi-object aggregation, which have been widely adopted as a baseline in a number of methods \cite{seong2020kernelized,hong2021adaptive,guo2022adaptive,wang2023look}.

Recently, some investigations demonstrate the advantages of integration modeling compared with highly discrete modeling. AOT \cite{yang2022associating} and DEAOT \cite{yang2022decoupling} (Fig. 1(b)) emphasize the shortage of single-object inference compared to joint modeling of multiple objects, particularly in exploring contextual and robust feature representations. Meanwhile, SimVOS \cite{wu2023scalable} (Fig. 1(c)) reveals limitations in the separation of feature extraction and matching processes for capturing dynamic object-aware features. However, existing methods solely focus on combining or aggregating substructures of the VOS pipeline, resulting in their suboptimal solutions. For example, AOT relies on sophisticated designing matching modules which fails to leverage superior pre-training weights, while the simplistic single-object framework of SimVOS encounters inefficiencies and the inapplicability of memory makes it unable to utilize rich features of the processed frames.

Therefore, we introduce an essentially novel framework distinct from previous methods, aiming to develop a unified end-to-end differentiable VOS model for global optimization. Based on this, the All-in-One Transformer for Video Object Segmentation (OneVOS) is proposed, which models and unifies all important modules into a Vision Transformer. Specifically, we first leverage the flexible attention mechanism to conduct feature extraction and mutual matching between the reference and current frames. To effectively capture multi-object relationships, we introduce a mask embedding layer that enriches features with multi-object modeling insights. Moreover, in the memory design, we propose an external layer-wise token memory to enhance the coherence understanding of videos by storing and providing spatio-temporal features for subsequent frames.

As the feature extraction and matching of multiple objects performing simultaneously, the reference tokens output from each layer of Transformer for memory update could be represented by a weighted linear combination of tokens from memory, reference frame and current frame, which may lead to a significant semantic errors and ambiguity to serving as features stored in memory. To address this, we further propose a Unidirectional Hybrid Attention (Uni-Hybrid Attention) through a double decoupling of the original attention operation. It guarantees the versatility of stored features and enhances memory performance. Notably, as the Uni-Hybrid Attention is derived from naive attention mechanism, our framework is still able to utilize pre-trained weights from various Vision Transformers, thereby enhancing its generality and flexibility.

 In this way, the proposed OneVOS framework implements end-to-end modeling and global optimization, exhibiting superior performance compared to previous models. To further strengthen its advantages, we conduct an in-depth exploration of the behavior within the attention layer of OneVOS to enhance its efficiency, which is particularly crucial in resource-constrained and long-term video tasks. Concretely, we introduce the Dynamic Token Selector (DTS), capable of dynamically selecting partial tokens from each layer for cross-attention and memory storage, trained solely under segmentation loss without additional constraints. With DTS integrated into every layer of the transformer, we analyze the token selection distribution of DTS and the attention weight distribution, uncovering a significant insight:  unlike the traditional sequential paradigm, OneVOS adopts a cross-layer segmentation mode that dynamically shifts focus between feature extraction and matching across different layers. Based on this, we propose a natural acceleration scheme involving designing distinct memory capacities for each layer and layer-wise memory updates via tokens selected by DTS. This approach enhances both speed and accuracy across multiple datasets.

We conducted comprehensive experiments across seven datasets, and OneVOS consistently achieves state-of-the-art performance across all of them. Notably, OneVOS demonstrates a substantial performance advantage enven in more challenging scenarios, achieving 71.1\% $\mathcal{J \& F}$ on LVOS validation and 67.2\% $\mathcal{J \& F}$ on MOSE validation. The main contributions are summarized as follows:
\begin{itemize}

\item We propose All-in-One Transformer framework for Video Object Segmentation (OneVOS), integrating feature extraction, matching, memory management, and objects aggregation into a vision transformer, which can be optimized globally. This provides a fundamentally novel paradigm in VOS.

\item We introduce the Unidirectional Hybrid Attention mechanism to facilitate multi-object feature extraction and matching, ensuring tokens are devoid of semantic ambiguity for efficient memory storage.

\item We propose the Dynamic Tokens Selector, which not only unveils the internal working mechanism but also leads to a more efficient OneVOS, based on the exploration and the resulting insight.

\item OneVOS achieves state-of-the-art performance across 7 datasets, particularly excelling in more complex LVOS and MOSE datasets with 70.1\% and 66.4\% $J \& F$, surpassing previous best methods by 4.2\% and 7.0\%, respectively.
\end{itemize}

\section{Related work}
\subsection{Video object segmentation}
Semi-supervised VOS methods can be divided into four types: 
The earlier fine-tuning methods 
\cite{caelles2017one,xiao2018monet,voigtlaender2017online,maninis2018video,bhat2020learning} learn general segmentation features via offline pre-training, then enhance object-specific representations by online fine-tuning. Then, propagation-based methods \cite{perazzi2017learning,oh2018fast,wang2018semi,johnander2019generative,cheng2018fast,hu2018motion} utilize optical flow to propagate the mask from the previous frame to the next one. Subsequent methods optimized it by keyframes enhancement \cite{khoreva2019lucid}, advanced tracking techniques \cite{cheng2018fast,xu2019mhp,huang2020fast,chen2020state}, and re-identification mechanisms \cite{li2018video}. Matching-based methods \cite{hu2018videomatch,chen2018blazingly,voigtlaender2019feelvos,yang2020collaborative,yang2021collaborative} identify the object of the current frame based on the correlation with the reference frame in the embedding space. 

Recently, Memory-based approaches \cite{seong2020kernelized,oh2019video,guo2022adaptive,hong2021adaptive} store features of segmented frames in a memory bank to provide a set of features for segmentation, which has become the dominant method in SVOS. Following that, methods such as HMMN \cite{seong2021hierarchical}, SSTVOS \cite{duke2021sstvos}, and XMem \cite{xmem} enhance performance by utilizing multi-scale features, introducing Transformer for matching, and building long-term memory. These methods typically adhere to a traditional discrete modeling and single-object architecture. More recently, the multi-object framework\cite{yang2022associating},\cite{yang2022decoupling} is proposed to address the inherent limitations of the previous single-object while still maintaining a sophisticated matching module for high performance. \cite{wu2023scalable}  presents a simplified VOS framework for joint feature extraction and matching which adopts single-object modeling and discards the vital memory design in VOS. Differently, our OneVOS, innovatively unifies feature extraction, matching, memory construction, and multi-objective aggregation all in one Transformer.  This design not only streamlines the segmentation process but also enhances the dynamic interaction among different stages of SVOS. 

\subsection{Visual Object Tracking}
Video object tracking (VOT) shares similarities with the task setting of SVOS, but it primarily utilizes bounding boxes for object localization instead of detailed segmentation masks. In VOT, there has been a paradigm shift from traditional Siamese-based methods towards leveraging Transformer layers for enriched relational modeling between the template and search regions. Pioneering works like TransT \cite{chen2021transformer} introduce a two-stream attention mechanism for robust bi-directional information exchange, while innovative methods like MixFormer \cite{cui2022mixformer} and AiATrack \cite{gao2022aiatrack} adopt asymmetric attention strategies to minimize distractions from irrelevant objects. Notably, OSTrack \cite{ye2022joint} proposes a unified attention model, promoting seamless feature integration and relation modeling. Nevertheless, the distinct multi-objects settings and difference in target information representation, coupled with the critical role of memory in VOS, render unification within the VOS field more challenging. And this paper endeavors to tackle this challenge.

\section{Method}

\begin{figure}[h]
  \centering
   \vspace{-5mm}
   \includegraphics[width=0.9\linewidth]{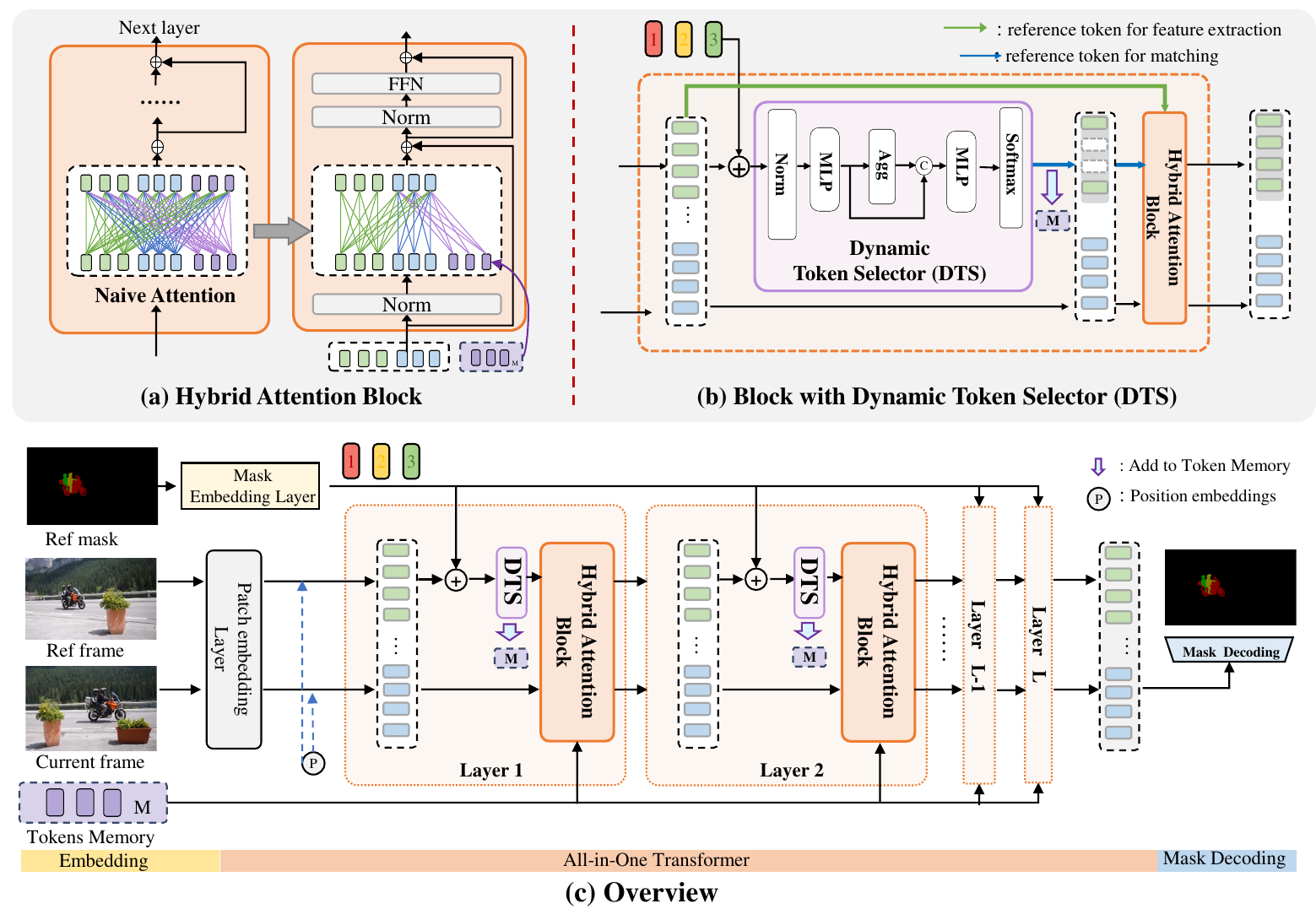}
   \vspace{-3mm}
   \caption{(a) Hybrid Attention Block. (b) Blocks with Dynamic Token Selector.  (c) Overview of OneVOS framework equipped with All-in-One Transformer. The reference mask is initially encoded into mask embeddings. Patch embeddings are generated from both the reference and current frames, which are then processed by the Transformer. Within each Transformer layer, mask embeddings are integrated with reference tokens. Then they are selected by Dynamic Token Selector and concatenated with tokens stored  in memory and current frame feeded into the Hybrid Attention Module. Finally, the features of current frame output from Transformer are subsequently decoded to multi-object masks.}
   \label{fig:overview}
   \vspace{-11mm}
\end{figure}

As shown in the Fig.~\ref{fig:overview}, in a given video sequence, VOS focuses on segmenting the current $t$-th frame $f_{t} \in \mathbb{R}^{H \times W \times 3}$ using the reference frame $f_{t-1} \in \mathbb{R}^{H \times W \times 3}$ with its mask and the history features stored in Memory. In this section, our aim is to construct an unified framework at a global scale to tackle the challenges of VOS. 
We commence by introducing the OneVOS framework in Sec. 3.1, which employs Vision Transformer to integrate core elements in VOS. And then Sec. 3.2 provides a detailed exposition on the novel Unidirectional Hybrid Attention mechanism. In addition, in Sec. 3.3, we explore the working mechanism of attention within  OneVOS through the proposed Dynamic Token Selector in order to introduce the efficient version of OneVOS.

\subsection{OneVOS Framework} 

\paragraph{\textbf{Mask Embedding.}}
 To capture complex target interactions and minimizing redundant single-target inferences, motivated by \cite{yang2022associating}, we utilize the Mask Embedding Layer (as Patch Embedding Layer in Transformer) to encode reference mask into Mask Embedding denoted as $m_{ref} \in \mathbb{R}^{ \frac{H}{P} \times \frac{W} {P} \times C}$ through a single convolutional layer. Here, $C$ represents the embedding dimension, with $P$ as the stride of the mask embedding. Utilizing the local relationship modeling capability of the convolution operation, the discriminability of multiple targets is enhanced, which will then integrated with visual feature map in Transformer before attention computation to integrating mask embeddings to the immediate value tokens of each layer, thereby effectively promoting the model to perform multiple targets inference.

\paragraph{\textbf{All-in-One Transformer.}}
Then, we adopt a transformer architecture to unify the core modules of VOS. It first generates the non-overlapping patch embeddings $e_{t}^0 \in \mathbb{R}^{N\times C}$ and $e_{ref}^0 \in \mathbb{R}^{ N\times C}$ for the current and reference frames, where $P$ is the patch resolution same as the mask embedding, thus $N=HW/P^2$. Then, the concatenation of $[e_{ref}^0;e_{t}^0]$ along the spatial dimension serves as the initial input for the backbone, facilitating adaptive multi-object feature extraction and target information propagation through the stacked layers. In each layer, we first integrate mask embedding with visual feature map as shown in Eq \ref{eq:add mask}:
\begin{equation}
\small
\begin{split}
\{Q^l_{ref}, K^l_{ref},{V^l_{ref}}\}= \{Q^l_{ref}, K^l_{ref},{V^l_{ref}+m_{ref}}\},
\end{split}
\label{eq:add mask}
\end{equation}
where $Q$, $K$, and $V$ are query, key, and value matrices aligning with the standard attention mechanism\cite{dosovitskiy2020image}, and $ l \in \{0,1,\cdots L\}$ represents the layer. Then, these object-aware reference tokens, memory tokens, and tokens of current frames undergo adaptive feature extraction and target information propagation by applying the attention mechanism of the Transformer. The output $a^l=[a^l_{ref};a^l_t]$ of layer $l$ after the attention operation is as follows:

\begin{equation}
\footnotesize
\begin{split}
\footnotesize
a^l  =  \mathrm{Softmax}(\frac{QK^{T}}{\sqrt{d_k}}) \cdot V  
 = \mathrm{Softmax} (\frac{[Q^l_{ref};Q^l_{t}] [K^l_{M};K^l_{ref};K^l_{t}]}{\sqrt{d_k}}) \cdot [V^l_{M};V^l_{ref};V^l_{t}],
\end{split}
  \label{eq:al}
\end{equation}
where $K^l_{M},V^l_{M}$ are features stored in memory. {Supposing $W=QK^T$ is the matrix of similarity weights, $a^l_{t}$ can be written as:}
\footnotesize
\begin{equation}
\footnotesize
a^l_{t}  = W_{t,M} V_{M} + W_{t,ref}V_{{ref}} + W_{t,t}V_{t},
  \label{eq:a}
\end{equation}
where  $[W_{t,ref}V_{ref}]$ and $[W_{t,M} V_{M}]$ are responsible for aggregating the inter-frame features from memory and reference frame to current frame (matching), and $[W_{t,t}V_{t}]$ aggregating the intra-frame features (feature extraction) for current frame, respectively.  Thus, dynamic coordination between feature extraction and matching is achieved through the attention weights learned adaptively. 

\paragraph{\textbf{Token Memory.}}
In addition, since the target information and the reference frame are integrated within our framework, we can store intermediate features directly into Token Memory without necessitating additional post-processing fusion steps. Given the semantic variance of intermediate tokens in the reference frame across layers, we introduce a token storage mechanism to save essential Transformer tokens instead of whole frames as follows:
\footnotesize
\begin{equation}
\footnotesize
M = \{K^l_{M},{V^l_{M}}  \ | \ \l \in \{0,1,\cdots L\}\} =\{K^l_{refs},{V^l_{refs}}  \ | \ \l \in \{0,1,\cdots L\}\},
  \label{eq:a}
\end{equation}
This method allows these tokens to be involved only in the attention modules insted multiple inference, greatly reducing inference overhead. 

\paragraph{\textbf{Mask Decoding.} }
Finally, the output of All-in-One Transformer, $e^L_t$ is input to the lightweight Feature Pyramid Network (FPN) for Mask Decoding. These features are then gradually upgraded to the original scale and decoded into multi-object masks.

\begin{figure*}[t]
  \centering
   \includegraphics[width=1\linewidth]{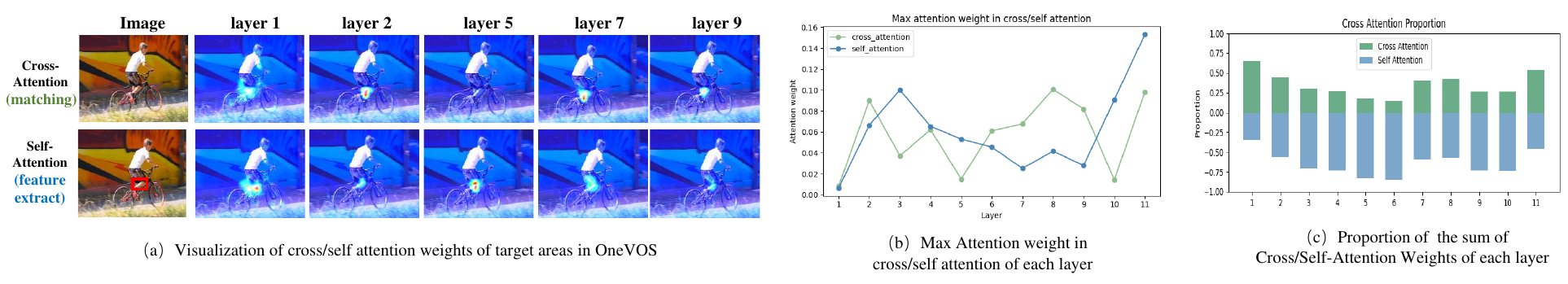}
\vspace{-5mm}
   \caption{(a) Visualization of attention distribution of target areas in OneVOS, showing the alternation between self-attention and cross-attention across layers. (b) Analysis of the maximal attention weights reveals a pattern of alternation between feature extraction via self-attention (notably in layers 3 and 5) and matching via cross-attention (as in layers 2 and 8). (c) The proportional analysis of the attention weights' sum, normalized to one, underscores the non-sequential prioritization across different layers between self-attention and cross-attention.}
   \label{fig:attenvis}
   \vspace{-5mm}
\end{figure*}

\subsection{Unidirectional Hybrid Attention}
Building upon our newly developed pipeline, all modules are integrated into a transformer as in Fig. \ref{fig:overview}(c). However, obtaining the stored features by employing original attention mechanism may lead to semantic errors and ambiguity since they are represented incorrectly. To address this challenge, our proposed Unidirectional Hybrid Attention (Uni-Hybrid Attention) is strategically designed through a dual-level attention decoupling as shown in Fig. \ref{fig:overview}(a).

Firstly, reference features calculated by naive attention operation to all the tokens are denoted as\ $a^l_{ref}=W_{ref,M} V_{M} + W_{ref,ref} V_{ref}+ W_{refs,t}V_{t}$. A key limitation in the computations arises where tokens of reference frame  containing target information (as in Eq. (1)) are represented by tokens of current frame  that without target information, causing semantic errors in the features stored in the memory which are essential for storing spatio-temporal target information for subsequent frames. Therefore, to alleviate this situation, our first modification decoupling attention from the current token to the reference token. The revised expression is:
\begin{equation}
\begin{small}
\begin{aligned}
a^l_{ref}=W_{ref,M} V_{M} + W_{ref,ref} V_{ref}.
\end{aligned}
\end{small}
\label{eq:decoup1}
\end{equation}

Subsequently,the attention computation between reference features and memory tokens is decoupled, serving a dual purpose: 1) Avoiding Information Overlap and Ambiguity: Memory tokens, enriched with mask embeddings, encapsulate target information from prior frames. Integrating these embeddings with those of the reference frame risks overlapping and confusing object cues. 2) Preserving Temporal Specificity: Every video frame contains essential temporal intricacies crucial for segmentation precision. Attention decoupling preserves the reference frame's temporal uniqueness, critical for accurate and context-aware segmentation. Therefore, we decouple attention between memory tokens and those from the reference frame, and the final unidirectional hybrid attention formulations are:
\begin{equation}
\begin{split}
& a^l_{ref}=  W_{ref,ref} V_{ref} \\
a^l_{t} &= W_{t,M} V_{M} + W_{t,ref} V_{ref}+ W_{t,t}V_{t}. \\
\end{split}
\label{eq:important}
\end{equation}

\subsection{Towards Efficient OneVOS} 
Utilizing the advanced All-in-One Transformer equipped with Uni-Hybrid Attention, OneVOS achieves outstanding performance across various datasets. However, the integration process of these modules in OneVOS stil lacks explainability, hindering the rational design of a more efficient version. Therefore, this section delves into its working mechanism to propose the efficient OneVOS.
\vspace{-2mm}

\begin{figure*}[t]
  \centering
     \includegraphics[width=1\linewidth]{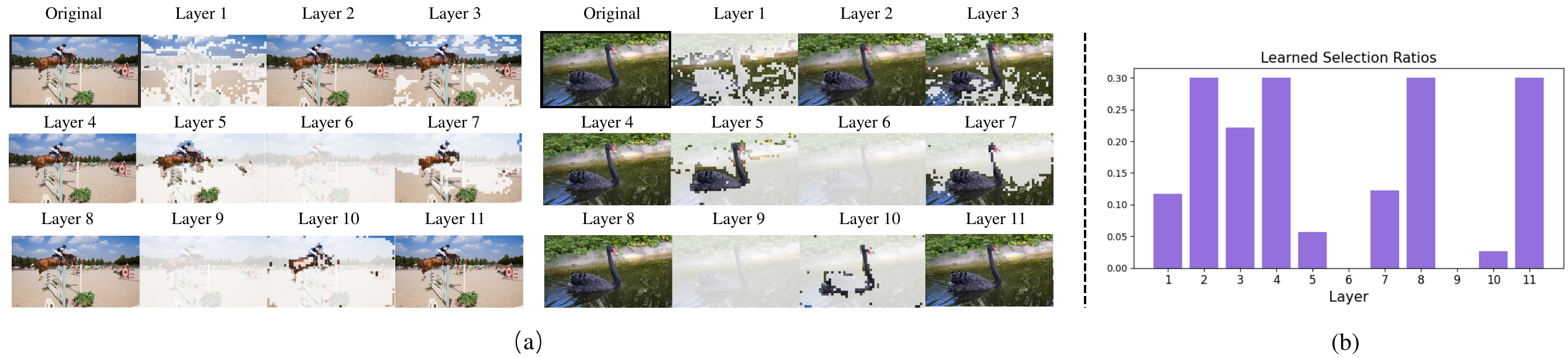}
   \vspace{-5mm}
   \caption{(a) Visualization of the selected tokens of reference frame by DTS in different Transformer layers. (b) Adaptive selection ratios learned by DTS in different layers without any constraints.}
   \label{fig:dts_selection}
   \vspace{-5mm}
\end{figure*}

\paragraph{\textbf{Exploration and Insight.}}
Within the Transformer architecture, attention weights determine the influence of each token in the input sequence on the output which is useful for interpreting Transformer behavior. By examining the attention weights of tokens in the target area of the current frame across each layer, we compare their focus in reference frames or themselves. As presented in Fig.  \ref{fig:attenvis}, qualitative results in \ref{fig:attenvis}(a) shown that a fluctuating focus between reference frames(cross-attention) and themselves(self-attention), Subsequently,  Fig. \ref{fig:attenvis} (b) and (c)  provide quantitative evidence of a complex alternation pattern, which deviates from the sequential feature extraction and matching process of traditional discrete VOS modeling or uniform distribution in each layer.  Instead, these findings lead us to a natural insight: OneVOS may learn a dynamic layer-wise integration strategy, preferentially executing multi-object feature extraction or matching in each layer according to the semantic context. \\
\vspace{-5mm}
\paragraph{\textbf{Dynamic Token Selector.}} 
Based on above insights and motivated by \cite{rao2021dynamicvit}, we introduce a Dynamic Token Selector (DTS) into the OneVOS architecture, as depicted in Fig. \ref{fig:overview}(b). This module adaptively selects a subset of tokens from the reference frame for cross-attention and storage.   The selection process, as defined in Eq. \ref{eq:DTS}, combines global context and local token information:
\begin{equation}
\begin{split}
& p = \mathrm{Softmax}(\mathrm{MLP}([\mathrm{MaxPool}(z^l_{ref});z^l_{ref}])) \\
& z^l_{ref} =  \mathrm{MLP}(\mathrm{Norm}(e^l_{ref}))
\end{split}
  \label{eq:DTS}
\end{equation}
where $z^l_{ref}$ denotes the local feature encoded of  each token, while $[MaxPool(z^l_{ref})]$ aggregates the overall context of the reference frame.  $p \in \mathbb{R}^{N \times 2}$ represents the probabilities associated with each token. Specifically, $p_{i,0}$ indicates the probability of dropping $i$-th token while $p_{i,1}$ indicates for selection. 

This design aims to enable the network to autonomously determine the optimal token selection ratio for high-performance segmentation. Consequently, it is integrated into an All-in-One Transformer, as depicted in Fig. \ref{fig:overview}(a). Moreover, it is exclusively trained under segmentation loss without additional constraints. By observing the proportion of tokens selected by DTS in each transformer layer, as illustrated in Fig. \ref{fig:dts_selection}(a) and Fig. \ref{fig:dts_selection}(b), a crucial insight emerges. The model automatically allocates varying selection ratios to different layers, and even assigns a selection ratio of zero or one to specific layers. This observation underscores the focus of DTS on certain layers for feature extraction (selecting a proportion of reference frame tokens as zero) and on specific layers for matching (selecting a proportion of reference frame tokens as one), presenting a characteristic of alternating focus across layers.

\begin{figure}[t]
  \centering
   \includegraphics[width=1\linewidth]{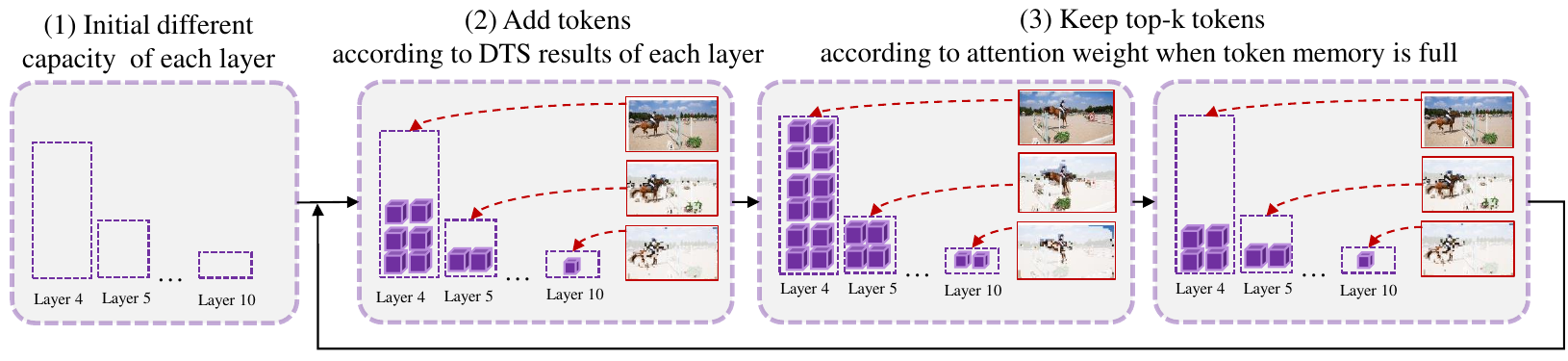}
\vspace{-5mm}
   \caption{ Visualization dynamic memory updates of OneVOS utilizing DTS. The selected tokens in red boxes are derived output from  DTS as Fig. \ref{fig:dts_selection}(a).}
   \label{fig:mem}
   \vspace{-5mm}
\end{figure}

\paragraph{\textbf{Efficient Version of OneVOS.}}
Building upon the above observations,  a straightforward acceleration method is naturally introduced. This includes layer-wise memory capacity and adaptive token update mechanism via DTS, as shown in Fig.~\ref{fig:mem}. Specifically, we  initialize different capacities for each layer of Token Memory, with the token capacity formula as follows:
\begin{equation}
c_l = r_l * N * Cap .
  \label{eq:dynamic}
\end{equation}
Here, $r_l$ represents the layer-specific selection ratio determined by DTS during training, as illustrated in Fig.  \ref{fig:dts_selection}(b). Additionally, $Cap$ is a hyper-parameter denoting the frame capacity (typically set to 3 or 5). Consequently, tokens are incrementally added to the memory based on these selection ratios until reaching the capacity limit of each layer. Subsequently, the top-k tokens, determined by the current attention weights $W_{t,ref}$, are retained, followed by the continuous addition of tokens to memory. Aligned with the segmentation mode of OneVOS, this method achieves improvements in accuracy and speed across multiple datasets.

\section{Experiment}
\begin{table*}[t]
\centering
\scriptsize 
\setlength{\tabcolsep}{1.1pt}
 \renewcommand{\arraystretch}{1}
\begin{tabular}{c|ccc|cccc|c|c|lcc}
\hline\hline
\multicolumn{8}{c|}{Long-term Videos} & \multicolumn{5}{c}{Complicated Videos} \\
\hline
\scriptsize 
\multirow{2}{*}{Method} & \multicolumn{3}{c|}{LVOS VAL} & \multicolumn{4}{c|}{LVOS TEST} & \multirow{2}{*}{Method} & \multirow{2}{*}{MOSE} & \multirow{2}{*}{$\mathcal{J \& F}$} & \multirow{2}{*}{$\mathcal{J}$} & \multirow{2}{*}{$\mathcal{F}$} \\
& $\mathcal{J \& F}$ & $\mathcal{J}$ & $\mathcal{F}$ & & $\mathcal{J \& F}$ & $\mathcal{J}$ & $\mathcal{F}$ & & & \\
\hline
CFBI \cite{yang2020collaborative} & 50.0 & 45.0 & 55.1 & & 44.8 & 40.5 & 49.0 & XMem  \cite{xmem} & \ding{56} & 56.4 & 52.1 & 60.6 \\
AOT-L \cite{yang2022associating}& 61.3  & 59.4 & 53.6 &  & 54.1 & 48.7 & 59.5 & DropSeg  \cite{wu2023dropmae}& \ding{56} & 52.2 & 48.0 & 56.4 \\
STCN \cite{cheng2021rethinking} & 45.8  & 41.1 & 50.5  &&  45.8 & 41.6 & 50.0 &R50-AOT-L\cite{yang2022associating} & \ding{56} & 58.1  & 53.9 & 62.3 \\
RDE  \cite{li2022recurrent}&52.9 & 47.7 & 58.1 & & 49.0 & 44.4 & 53.5 & R50-DeAOT-L \cite{yang2022decoupling}& \ding{56}  & 58.9& 54.5 & 63.3 \\
XMem  \cite{xmem}& 50.0 & 45.5 & 54.4 & & 49.5 & 45.2 & 53.7 & SwinB-AOT-L \cite{yang2022associating} & \ding{56} & 60.0  & 56.1 & 64.0 \\
AOT-B  \cite{yang2022associating}  & 56.9 & 51.8 & 61.9 & & 54.4 & 49.3 & 59.4 &  \textbf{OneVOS} & \ding{56} &\textbf{62.3}  & \textbf{58.1} & \textbf{66.6} \\
DeAOT  \cite{yang2022decoupling}&63.7 & 57.6 & 69.8 & &58.1 & 52.9 & 63.4  & RDE\cite{li2022recurrent} & \ding{52}& 48.8 & 44.6 & 52.9 \\
R50-AOT-L \cite{yang2022associating} & 63.8 & 58.1 & 69.6 &  & 61.0 & 56.1 & 65.9 & STCN\cite{cheng2021rethinking} & \ding{52} & 50.8  & 46.6  & 55.0 \\
R50-DeAOT-L \cite{yang2022decoupling} &  65.9 & 59.9 & 71.9  & & 58.0 & 53.2 & 62.8 &  XMem \cite{xmem}& \ding{52} &57.6& 53.3 & 62.0 \\
Swin-AOT-L  \cite{yang2022associating}& 65.6 & 60.8 & 75.1 & & 61.5 & 56.4 & 74.8 & DeAOT  \cite{yang2022decoupling}& \ding{52} & 59.4  & 55.1  & 63.8 \\
\hline
\textbf{OneVOS-D}  & \underline{70.1} &\underline{64.7} & \underline{75.5} & & 60.1 & 55.6 & 68.0  &\textbf{OneVOS-D} & \ding{52} &61.7  & 57.5 & 66.0 \\
\textbf{OneVOS}  & 67.4 & 62.0 & 72.9 & & \underline{63.8} & \underline{59.5} & \textbf{70.0} & \textbf{OneVOS} & \ding{52} &\underline{66.4}  & \underline{62.2} & \underline{70.6}\\
\textbf{OneVOS*} & \textbf{71.1} & \textbf{66.0} & \textbf{76.2} & &  \textbf{64.1} & \textbf{60.0} & \underline{68.2} & \textbf{OneVOS*} & \ding{52} & \textbf{67.2} & \textbf{63.0} & \textbf{71.5} \\
\hline\hline
\end{tabular}
\vspace{+1mm}
\caption{Quantitative comparison of long-duration video datasets (LVOS validation and test) and complex datasets (MOSE). \ding{52} indicates uses MOSE for main stage training.}
\label{compliate_comp}
\vspace{-8mm}
\end{table*}

\subsection{Implementation Details}
One-stream VOS uses ConvVIT-base \cite{gao2022convmae} as the backbone, initialized with ImageNet-1k\cite{deng2009imagenet} pre-training weights. The feature resolution of the mask decoding is increased from 1/16 to 1/4 and the channel dimension is reduced from 768 to 384. 
\vspace{-2mm}
\paragraph{\textbf{Training details.}}
Following the common settings of VOS\cite{cheng2021modular,yang2022associating}, the training phase is divided into two stages: pre-training on synthetic video sequences generated from still image datasets \cite{everingham2010pascal,lin2014microsoft,cheng2014global,shi2015hierarchical,hariharan2011semantic}, and main training on standard VOS dataset (DAVIS 17 and YouTube VOS in default, the model mark with "*" uses MOSE in main training as well). The loss function is a 0.5:0.5 combination of bootstrapped cross-entropy loss and soft Jaccard loss\cite{nowozin2014optimal}. The learning rate of two stages is 2e-4 and 1e-4, respectively, utilizing  4 $\times$ 3090s GPUs.  
 Our batchsize is 4, with a gradient accumulation of 2 and a sequence length of 5, running each training phase for 200,000 iterations. In the training of DTS, we employ the Gumbel-Softmax\cite{jang2016categorical} on its output $p$ to generate a one-hot selection vector, indicating whether each token is selected (assigned a value of 1) or not (value of 0). Throughout training, we record the selection rate, observing that in the later stages of training, the selection rate of each layer gradually stabilizes.  
 \vspace{-3mm}
\paragraph{\textbf{Inference details.} }
OneVOS utilizes basic memory storage, storing tokens at intervals of 5 or 10 frames and updates using the first-in-first-out queue when it is full, except for the initial frame tokens. Conversely, OneVOS-D employs dynamic memory, defaulting to retain a maximum of 520 top-k tokens when full. The maximum frame-level capacity $Cap$ varies by dataset, set to 3/5/8. For fair comparison in LVOS, we strictly limit memory to under 6.  
\vspace{-3mm}
\paragraph{\textbf{Datasets and Metrics.}} We evaluate the performance of OneVOS on 4 standard VOS datasets, including  DAVIS 16 val \cite{perazzi2016benchmark}, DAVIS 17 val \cite{pont20172017}, DAVIS 17 test \cite{pont20172017}, and YouTube 2019 \cite{xu2018youtube}, as well as 3 more challenging datasets,  MOSE \cite{ding2023mose}, LVOS val, and LVOS test \cite{hong2022lvos}. The evaluation metrics include region similarity $\mathcal{J}$, boundary accuracy $\mathcal{F}$, and their average $\mathcal{J \& F}$, and FPS for speed. For YouTube-VOS, we also report them from the seen $\mathcal{J}_s$, $\mathcal{F}_s$  and unseen $\mathcal{J}_u$, $\mathcal{F}_u$ categories.

\begin{table*}[t]
  \centering
  \setlength{\tabcolsep}{1pt}
  \renewcommand{\arraystretch}{1}
  \scriptsize 
  \begin{tabular}{c|ccc|cccc|ccc|ccccc}
  \hline
  \hline
  {\multirow{2}{*}{Method}}  & \multicolumn{3}{c|}{{DAVIS 16 VAL}}  & \multicolumn{4}{c|}{{DAVIS 17 VAL}}   & \multicolumn{3}{c|}{{DAVIS 17 TEST}} & \multicolumn{5}{c}{{YOUTUBE 19 VAL}} \\
  {} & $\mathcal{J \& F}$ & $\mathcal{J}$& $\mathcal{ F}$ & $\mathcal{J \& F}$ & $\mathcal{J}$ & $\mathcal{ F}$ & $\mathcal{FPS}$
  & 
  $\mathcal{J \& F}$ & $\mathcal{J}$ & $\mathcal{F}$ 
  &
  $\mathcal{G}$ & $\mathcal{J}_{s}$ & $\mathcal{F}_s$ & $\mathcal{J}_{u}$ & $\mathcal{F}_{u}$ \\
  \hline
  CFBI+  \cite{yang2021collaborative}& 89.9 & 88.7 & 91.1 & 82.9 & 80.1 & 85.7 & 5.6 & 78.0 & 74.4 & 81.6 & 82.6 & 81.7 & 86.2 & 77.1& 85.2\\
  HMMN  \cite{seong2021hierarchical}& 90.8 & 89.6 & 92.0 & 84.7 & 81.9 & 87.5 & 5.0 & 78.6 & 74.7 & 82.5 & 82.5 & 81.7 & 86.1 & 77.3 & 85.0\\
   DropSeg  \cite{wu2023dropmae}& 92.1 & 90.9 & 93.3 & 86.5 & 83.5 & 89.5 & - & - & - & - & 83.4 & 82.9 & 87.3 & 77.7 & 85.6 \\
  STCN   \cite{cheng2021rethinking}& 91.6 & 90.8 & 92.5 & 85.4 & 82.2 & 88.6 & 20.2 & 76.1 & 72.7 & 79.6 & 82.7 & 81.1 & 85.4 & 78.2 & 85.9 \\
  XMem  \cite{xmem}&  91.5 & 90.4 & 92.7 & 86.2 & 82.9 & 89.5 & 22.6 & 81.0 & 77.4 & 84.5 & 85.5 & 84.3 & 88.6 & 80.3 & 88.6 \\
  Swin-AOTL\cite{yang2022associating}& 92.0 & 90.7 & 93.3 & 85.4 & 82.4 & 88.4 & 12.1 & 81.2 & 77.3 & 85.1 & 84.5 & 84.0 & 88.8 &78.4 & 86.7 \\
  Swin-DeAOTL \cite{yang2022decoupling}& 92.9 & 91.1 & 94.7 & 86.2 & 83.1 & 89.2 & 15.4 & 82.8 & 78.9 & 86.7 & {86.1} & \textbf{85.3} &  \textbf{90.2} & 80.4 & 88.6\\
  ISVOS  \cite{wang2023look}&  92.6 & 91.5 &  93.7 & 87.1 & 83.7 & 90.5 & - & 82.8 & 79.3 & 86.2 & {86.1} &{85.2} & 89.7 & 80.7 & 88.9 \\
  \cline{1-16}
MiVOS' \cite{cheng2021modular} & 91.0 & 89.6 & 92.4 & 84.5 & 81.7 & 87.4 & 11.2 & 78.6 & 74.9 & 82.2 & 82.4 & 80.6 & 84.7& 78.1 &86.4 \\
  STCN' \cite{cheng2021rethinking} & 91.7 & 90.4 & 93.0 & 85.3 & 82.0 & 88.6 & 20.2 & 77.8 & 74.3 &81.3 & 84.2& 82.6 &87.0 &79.4 &87.7 \\
  XMem' \cite{xmem} & 92.0 & 90.7 & 93.2 & 87.7 & 84.0 & 91.4 & \textbf{22.6} & 81.2 & 77.6 & 84.7 & 85.8 & 84.8 &89.2& 80.3 &88.8 \\
   SimVOS \cite{wu2023scalable} & 92.9 & 91.3 & 94.4 & 88.0 & 85.0 & 91.0 & 4.9 & 80.4 & 76.1 & 84.6 & 84.2 & 83.1 &-&79.1 &- \\
  \cline{1-16}
 \textbf {OneVOS} & 92.7	& 91.0 & 94.3 & {88.5} & {84.6} & \textbf{92.4} & 12.0 & {84.8} & {80.7} & {88.9} & {86.1} & {84.9 }& {89.8} & {80.6} & {89.1} \\
  \textbf{OneVOS-D} & \textbf{93.1} & \textbf{91.4} & \textbf{94.8} & \textbf{88.8} & \textbf{85.2} & \textbf{92.4} & 16.0 & 83.9 & 80.0 & 87.6 & 85.0 & 84.5 & 89.2 & 79.1 & 87.3 \\
  \textbf{OneVOS*} & {92.8} & {91.0} & {93.7} & 88.3 & 84.4 & {92.1} & 12.0 & \textbf{85.3} & \textbf{81.3} & \textbf{89.3} & \textbf{86.3} & 85.0 & {89.9} & \textbf{81.2} & \textbf{89.1} \\ 
  \hline
  \hline
  \end{tabular}
  \vspace{+1mm}
   \caption{Quantitative comparisons on the DAVIS 2016 val, DAVIS 2017 val, and YouTube-VOS 2019 val split. The methods Use ' as a superscript indicates adding the BL30k \cite{cheng2021modular} dataset for training.}
  \label{tab:davis_ytb}
  \vspace{-5mm}
\end{table*}
 \begin{figure*}[t]
  \centering
    \includegraphics[width=1\linewidth]{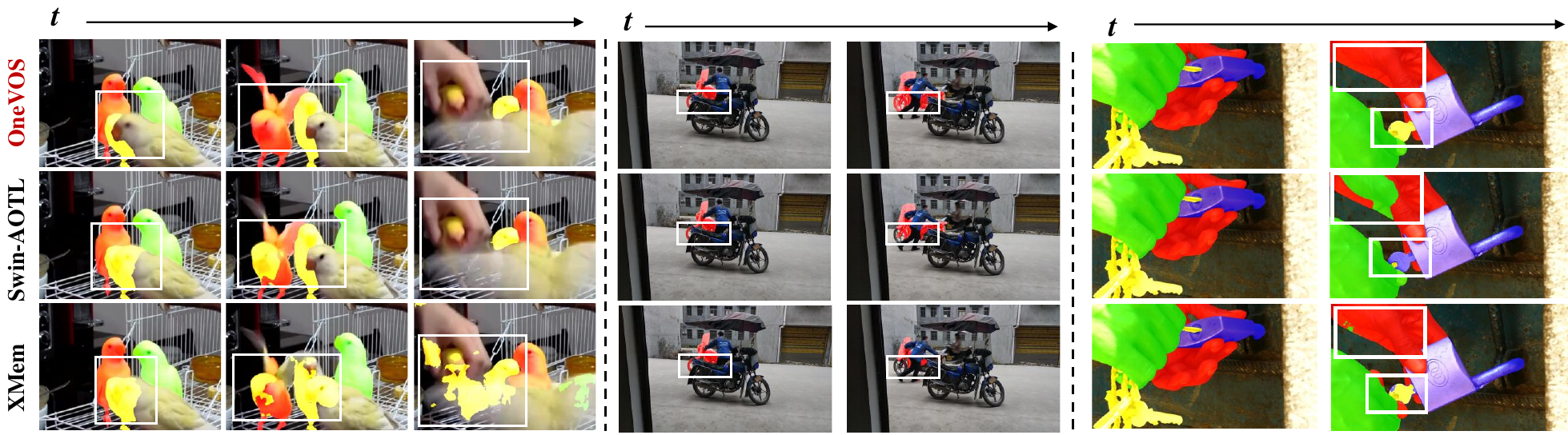}
   \caption{Qualitative comparisons between the state-of-the-art models XMem and Swin-AOTL in challenging short-term videos.}
   \label{fig:viscompare}
   \vspace{-5mm}
\end{figure*}

\subsection{Comparison with State-of-the-art Methods}
\begin{figure*}[t]
  \centering
   \includegraphics[width=1\linewidth]{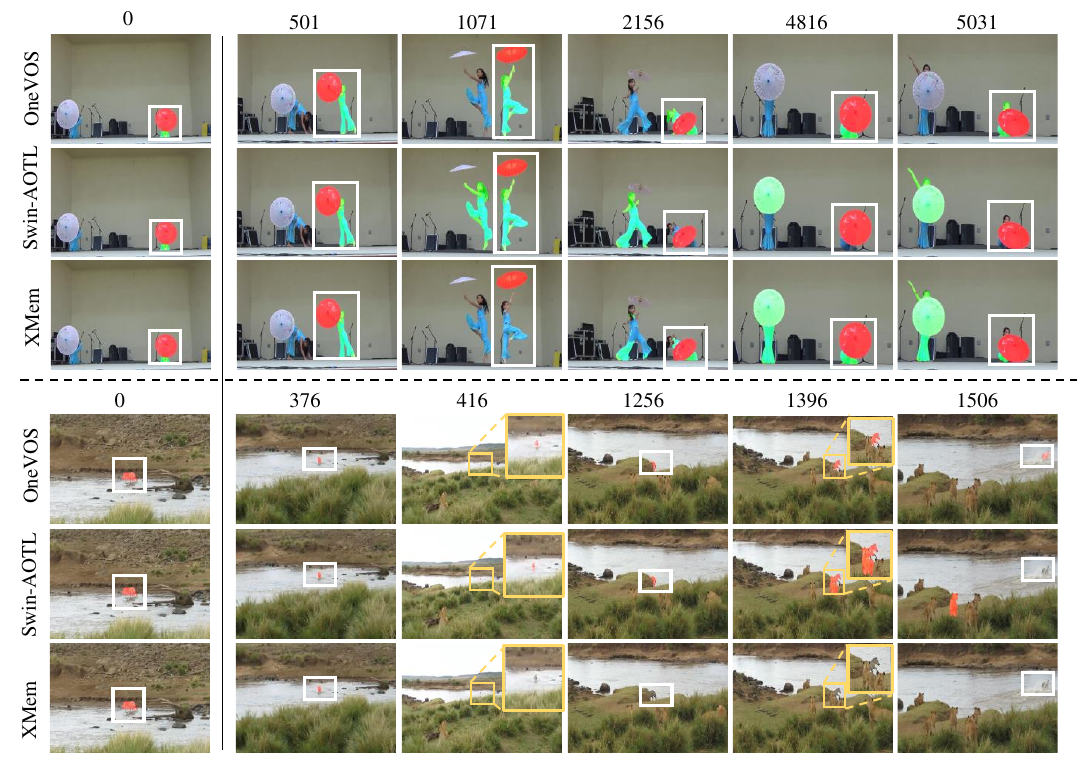}
   \caption{Qualitative comparisons between  the state-of-the-art models XMem and Swin-AOTL in challenging long-term videos.}
   \label{fig:viscompare_long}
    \vspace{-5mm}
\end{figure*}

\paragraph{\textbf{Complicated datasets.}}
Table \ref{compliate_comp} tabulates the quantitative results on the LVOS val, LVOS test, and MOSE. OneVOS demonstrates superior performance across multiple settings and datasets. Notably, OneVOS-D excels in the long video dataset LVOS, achieving an impressive 70.1\% $J \& F$ score under identical training conditions, demonstrating enhanced robustness in long-term videos.  On the MOSE dataset, results of OneVOS are particularly noteworthy, surpassing the highest results reported in the paper by nearly 7\% (66.4\% vs 59.4\%) trained in the same setting. When trained on YouTube, DAVIS, and MOSE together in the main training, OneVOS* achieves state-of-the-art performance across all three datasets. 
\paragraph{\textbf{Common datasets.}}
Table \ref{tab:davis_ytb} tabulates the quantitative results on the four standard datasets. The results clearly highlight the superiority of OneVOS. OneVOS-D achieves the best 93.1\% in DAVIS 2016 val and 88.8\% in DAVIS 2017 val respectively,  setting new state-of-the-art performance. OneVOS and OneVOS* showcase remark improvement in DAVIS 2017 test and YouTube-VOS 2019 val. This proves that OneVOS has excellent segmentation capabilities in both regular and challenging scenarios, demonstrating its practicability. 
\paragraph{\textbf{Qualitative Results.}}  We present  visualization results of OneVOS comparing against the state-of-the-art  method XMem \cite{xmem} and Swin-AOTL \cite{li2022recurrent}. The results for the short-term datasets and complicated long-term videos are shown in Fig. \ref{fig:viscompare} and Fig. \ref{fig:viscompare_long}, respectively. OneVOS consistently achieves finer and higher-quality segmentation on the short-term datasets. Notably, it showcases superior segmentation capabilities in long-term scenarios involving both background interference and small targets.\\

 \vspace{-5mm}
\subsection{Ablation Study}
In this section, we conduct a comprehensive analysis of the OneVOS framework and the efficacy of its core designs on both common and the more challenging datasets.

Table \ref{tab:ablation_study} details an incremental ablation study highlighting the significance of key design elements in OneVOS. Starting with a baseline where ConvVIT replaces the backbone in a top discrete multi-model framework \cite{yang2022associating} (line 1). Furthermore, to negate any bias stemming from varying modeling approaches toward the backbone, we employ an optimal model based on Swin Transformer\cite{liu2021swin}  as an advanced baseline (line 2), providing a stringent benchmark to validate the effectiveness of OneVOS. 
 \vspace{-3mm}
\paragraph{\textbf{All-in-One Modeling.}} We initially evaluate the efficacy of All-in-One Modeling by analyzing lines 1, 2, and 3 of Table \ref{tab:ablation_study}. The outcomes highlight the substantial benefits of {All-in-One Modeling. Our model registers a 2.45\% improvement in performance on DAVIS 17 over the enhanced baseline. Moreover, in long video datasets like LVOS, even without any specialized module designs and using only the basic self-attention mechanism, our model exhibits only a minor decrease in performance, demonstrating robust performance of Unification modeling on complex long video datasets. 
 \vspace{-2mm}
\paragraph{\textbf{Unidirectional Hybrid Attention.}} In our unidirectional hybrid attention module analysis, shown in lines 3 to 6 of Table \ref{tab:ablation_study}. Notably,  through an examination of lines 4 and 6 we observed that the integration of dual decoupling significantly enhances model performance on both DAVIS 17 val and LVOS datasets. However, Sole decoupling from the current to reference frame results in a performance drop. We analyze that the attributed to the increased focus of reference frame on memory tokens, thereby amplifying the ambiguity in mask embedding and hindering accurate target identification. On the other hand, decoupling solely from memory tokens to reference frame yields a marginal performance enhancement in DAVIS 17 val. However, this improvement is limited as the target features with objects information are represents by tokens from current frame without targets information, failing to provide a consistent feature representation across the video sequence so the performance drops in long videos. These findings highlight the necessity of dual decoupling is indispensable, and their joint effect promotes optimal model performance.

\begin{table*}[t]
\scriptsize
\centering
\setlength{\tabcolsep}{3pt}
\renewcommand{\arraystretch}{1.05}
\begin{tabular}{c|c|c|c|c|ccc|ccc}
\hline
\hline
\multirow{2}{*}{Method} & \multirow{2}{*}{All-in-One} & \multirow{2}{*}{Decoup1} & \multirow{2}{*}{Decoup2} & \multirow{2}{*}{DTS} &\multicolumn{3}{c|}{DAVIS 17 VAL} & \multicolumn{3}{c}{LVOS VAL} \\
& & & & & $\mathcal{J \& F}$ & $\mathcal{J}$ & $\mathcal{F}$  & $\mathcal{J \& F}$ & $\mathcal{J}$ & $\mathcal{F}$ \\
\hline
{1} & & & & & 82.8 & 79.7 & 85.9 & 60.8 & 56.4 & 65.1  \\
{2} & & & & & 85.4 & 82.4 & 88.4 & 65.6 & 60.8 & 75.1 \\
\hline
\textcolor{red}{3} & \ding{52} & & & & 87.5 & 84.3 & 90.7 & 64.9 & 60.3 & 69.5 \\
\textcolor{red}{4} & \ding{52} & \ding{52} & & & 87.1 & 84.0 & 90.1 & 61.0 & 55.7 & 66.2 \\
\textcolor{red}{5} & \ding{52} & & \ding{52} & & 87.6 & 84.4 & 90.8& 62.3& 57.1 &67.5  \\
\textcolor{red}{6} & \ding{52} & \ding{52} & \ding{52} & & 88.5 & 84.6 & 92.4& 67.4 & 62.0 & 72.9 \\
\textcolor{red}{7} & \ding{52} & \ding{52} & \ding{52} & \ding{52} & \textbf{88.8} & \textbf{85.2} & \textbf{92.4}  & \textbf{70.1} & \textbf{64.7}  & \textbf{75.5} \\
\hline
\hline
\end{tabular}
 \vspace{+1mm}
\captionof{table}{Ablation on Core Design of OneVOS. (1-2) baseline: multi-object discrete method with ConvVIT or Swin Transformer\cite{liu2021swin} as backbone. (\textcolor{red}{3-7}): OneVOS with ConvVIT: 'All-in-One': All-in-One modeling. 'Decoup1': decouple attention from current to reference, 'Decoup2': decouple attention between memory and reference. 'DTS': DTS with dynamic memory.}
\label{tab:ablation_study}
\vspace{-3mm}
\end{table*}
\begin{table*}[t]
\footnotesize
\begin{subtable}{.5\linewidth}
\hfill
 \setlength{\tabcolsep}{1.1pt}
 \renewcommand{\arraystretch}{1.3}
\centering
\scriptsize 
 \begin{tabular}{c|lcc|lcc}
  \hline
   \hline
    {\multirow{2}{*}{Method}} & \multicolumn{3}{c|}{{DAVIS17  VAL}}  & \multicolumn{3}{c}{{MOSE}} \\
    {} & $\mathcal{J \& F}$ & $\mathcal{J}$& $\mathcal{ F}$ &  $\mathcal{J \& F}$ & $\mathcal{J}$& $\mathcal{ F}$  \\ 
  \cline{1-7}
     first 6 layers & 87.6 & 83.7 & 91.4 & 60.9 & 56.8 & 65.1 \\
last 6 layers& 88.0 & 84.4 & 91.6 & 59.1 & 55.1 & 63.1 \\
 all layers & \textbf{88.5} & \textbf{84.6} & \textbf{92.7} &\textbf{62.3}  & \textbf{58.1} & \textbf{66.6} \\
\hline
    \hline
  \end{tabular}
   \label{tab:add_layers}
\end{subtable}%
\begin{subtable}{.5\linewidth}
\scriptsize 
 \setlength{\tabcolsep}{1.1pt}
 \renewcommand{\arraystretch}{1.3}
\centering
\scriptsize 
\begin{tabular}{c|ccc|ccc}
\hline
\hline
{\multirow{2}{*}{Method}} & \multicolumn{3}{c|}{DAVIS 17 VAL} & \multicolumn{3}{c}{LVOS VAL} \\
{} & $\mathcal{J \& F}$ & $\mathcal{J}$ & $\mathcal{ F}$ & $\mathcal{J \& F}$ & $\mathcal{J}$ & $\mathcal{ F}$ \\
\hline
DTS-t & 87.2 & 83.6 & 90.8 & 66.4 & 61.2 & 71.6 \\
DTS-$L_{r}$ & 86.7 & 83.1 & 90.3 & 60.2 & 54.9 & 65.5 \\
DTS & \textbf{88.8} & \textbf{85.2} & \textbf{92.4} & \textbf{70.1} & \textbf{64.7} & \textbf{75.5} \\
\hline
\hline
\end{tabular}
  \label{tab:keep_token}
\end{subtable}
 \vspace{+1mm}
\caption{From left to right: (a) Ablations on different mask embedding adding layers and (b) Ablation of different DTS designs in OneVOS.}
\label{tab:ab_addlayer_dts}
 \vspace{-7mm}
\end{table*} 

\paragraph{\textbf{Dynamic Tokens Selector.}} The comparison results between lines 6 and 7 of Table \ref{tab:ablation_study} highlights the role of DTS. DTS not only probes the segmentation mechanisms within OneVOS but, when integrated into the network, it bolsters the learning of this paradigm. It contributes to a more efficient OneVOS by adaptively selecting tokens that align with the model's feature extraction and matching mechanisms, culminating in state-of-the-art performance on both datasets.
\paragraph{\textbf{Mask Embedding Add Layers.} }  Table. \ref{tab:ab_addlayer_dts}(a) explores the effect of integrating mask embedding at different transformer layers. Results show that, by integrating mask embeddings to the immediate value tokens of each layer, the model can capture the global and local semantics of targets at different levels, thereby effectively promoting the model to perform multiple targets inference. Notably, embedding in the first six layers proved more beneficial than in the latter six, this indicates that early interaction between the current and reference frames without target information may negatively impact the feature extraction of the current frame.
\paragraph{\textbf{DTS design.}} The role of DTS is to judiciously select tokens by amalgamating each token feature of the reference frame with the global features of all tokens. Table \ref{tab:ab_addlayer_dts}(b) explores the effect of various DTS configurations, including the method using pooled tokens from the current frame as global information (DTS-t), revealing that guiding selection with the current frame may hinder the generalization of OneVOS across frames, consistent with our initial attention decoupling. Besides, Implementing a loss function to maintain a fixed 0.5 selection rate per layer (DTS-$L_{r}$) significantly reduces performance, underscoring that rigid selection criteria detrimentally affect the capabilities of OneVOS for adaptive matching and feature extraction. 

\begin{table*}[t]
\centering
\scriptsize
\begin{subtable}{.37\linewidth}
\centering
\setlength{\tabcolsep}{1.8pt}
\renewcommand{\arraystretch}{1.1}
\begin{tabular}{ccccc}
\hline
\hline
$Cap$ & D17-V & D17-T & FPS & Mem(G) \\
\hline
1 & 87.5 & 83.2 & \textbf{16.6} & \textbf{1.72} \\
2 & 88.0 & 84.4 & 13.7 & 2.20 \\
3 & 88.4 & 84.6 & 12.9 & 2.47 \\
4 & \textbf{88.5} & \textbf{84.6} & 12.3 & 2.69 \\
5 & 88.5 & 84.8 & 12.0 & 2.94 \\
\hline
\hline
\end{tabular}
\label{tab:add_layers}
\end{subtable}%
\begin{subtable}{.29\linewidth}
\centering
\setlength{\tabcolsep}{1.8pt}
\renewcommand{\arraystretch}{1.3}
\begin{tabular}{c|c|c}
\hline
\hline
{\multirow{2}{*}{Tokens}} & \multicolumn{1}{c|}{{MOSE}}  & \multicolumn{1}{c}{{LVOS VAL}} \\
{} & $\mathcal{J \& F}$ &  $\mathcal{J \& F}$  \\ 
\hline
256 & 59.6  & 69.3 \\
512 & 60.6 & \textbf{70.1} \\
1024 & \textbf{60.8} & 68.7 \\
\hline
\hline
\end{tabular}
\label{tab:keep_token}
\end{subtable}%
\begin{subtable}{.33\linewidth}
\centering
\setlength{\tabcolsep}{1.8pt}
\renewcommand{\arraystretch}{1.3}
\begin{tabular}{c|lcc}
\hline
\hline
\multirow{2}{*}{Backbone} & \multicolumn{3}{c}{DAVIS17  VAL} \\
& $\mathcal{J \& F}$ & $\mathcal{J}$ & $\mathcal{ F}$ \\
\hline
VIT & 87.5 & 84.1  & 91.0 \\
ConvVIT-b & 88.5 & 84.6 & 92.7 \\
ConvVIT-l & \textbf{88.9} & \textbf{85.3} & \textbf{92.4} \\
\hline
\hline
\end{tabular}
\label{tab:backbone_comp}
\end{subtable}
\vspace{+1mm}
\caption{Ablations on 3 specific module design of All-in-One framework. From left to right: (a) different $Cap$ of dynamic memory, where D17-V and D17-T denote the ${J \& F}$ metris of DAVIS-17 VAL and TEST datasets, respectively. (b) Different top-k tokens for dynamic memory, and (c) Different Transformer backbones of OneVOS.}
\label{tab:ablation3}
\vspace{-8mm}
\end{table*}

\paragraph{\textbf{Dynamic Memory Capacity.}} Various memory capacities (${Cap}$ in Eq. \ref{eq:dynamic}) impact accuracy, inference speed, and storage use. The performance of OneVOS under different memory capacities is validated in Table \ref{tab:ablation3}(a). The results reveal that while memory plays a crucial role in handling complex videos, OneVOS achieves robust performance even with just two frames or a single frame in common scenarios, thanks to its unified framework and efficient architecture. Optimal performance is observed when the memory capacity reaches 4 or 5. This underscores the importance of memory in enabling OneVOS to effectively handle longer and more intricate videos. 
\paragraph{\textbf{Number of Retained Nodes.}}  Table. \ref{tab:ablation3}(b) focuses on retaining a set number of top-k nodes in token memory, based on attention weights. Results show that after reaching a per-layer threshold. keeping approximately 512 nodes struck an optimal balance between information depth and computational efficiency.
\paragraph{\textbf{Backbone.}} In Table. \ref{tab:ablation3}(c), we assess Vision Transformers (ViT)\cite{dosovitskiy2020image} pre-trained with Masked Autoencoders (MAE) \cite{he2022masked}  and ConvVIT (base and large variants) as backbones for OneVOS. With the VIT-B backbone OneVOS still outperforms  most of the methods on D17 datasets. However, a key limitation of ViT was identified in its direct 16x downsampling approach, which resulted in a significant loss of multi-scale detail information. This is particularly detrimental for high-precision, pixel-level VOS tasks.  On the other hand, the ConvVIT model, which employs simple convolutional operations for its patch embedding process, was found to effectively preserve multi-scale information. Additionally, the increased complexity of the ConvVIT-large variant further enhanced the model's performance, underscoring the importance of architectural design in capturing intricate details necessary for accurate VOS.

\section{Conclusion}
This paper introduces OneVOS, a novel semi-supervised video object segmentation framework that globally integrates feature extraction, matching, memory management, and object aggregation within a single Transformer architecture. By employing the All-in-One Transformer with specialized Unidirectional Hybrid Attention modules, we significantly enhance the practicality and performance of OneVOS. Furthermore,  to propose a rational, efficient version of OneVOS, we investigate the mechanism of the All-in-One Transformer via the proposed Dynamic Token Selector, revealing its layer-crossing segmentation mode for multi-object feature extraction and matching, leading to a naturally efficient version with dynamic memory. OneVOS demonstrates state-of-the-art performance on 7 datasets, particularly excelling in complex scenarios. We anticipate that this work will draw greater attention to the unified VOS modeling.

\clearpage  

%
%
\bibliographystyle{splncs04}
\bibliography{main}

\begin{thebibliography}{10}
\providecommand{\url}[1]{\texttt{#1}}
\providecommand{\urlprefix}{URL }
\providecommand{\doi}[1]{https://doi.org/#1}

\bibitem{bhat2020learning}
Bhat, G., Lawin, F.J., Danelljan, M., Robinson, A., Felsberg, M., Van~Gool, L., Timofte, R.: Learning what to learn for video object segmentation. In: Computer Vision--ECCV 2020: 16th European Conference, Glasgow, UK, August 23--28, 2020, Proceedings, Part II 16. pp. 777--794. Springer (2020)

\bibitem{caelles2017one}
Caelles, S., Maninis, K.K., Pont-Tuset, J., Leal-Taix{\'e}, L., Cremers, D., Van~Gool, L.: One-shot video object segmentation. In: Proceedings of the IEEE conference on computer vision and pattern recognition. pp. 221--230 (2017)

\bibitem{chen2020state}
Chen, X., Li, Z., Yuan, Y., Yu, G., Shen, J., Qi, D.: State-aware tracker for real-time video object segmentation. In: Proceedings of the IEEE/CVF Conference on Computer Vision and Pattern Recognition. pp. 9384--9393 (2020)

\bibitem{chen2021transformer}
Chen, X., Yan, B., Zhu, J., Wang, D., Yang, X., Lu, H.: Transformer tracking. In: Proceedings of the IEEE/CVF conference on computer vision and pattern recognition. pp. 8126--8135 (2021)

\bibitem{chen2018blazingly}
Chen, Y., Pont-Tuset, J., Montes, A., Van~Gool, L.: Blazingly fast video object segmentation with pixel-wise metric learning. In: Proceedings of the IEEE conference on computer vision and pattern recognition. pp. 1189--1198 (2018)

\bibitem{xmem}
Cheng, H.K., Schwing, A.G.: Xmem: Long-term video object segmentation with an atkinson-shiffrin memory model. In: European Conference on Computer Vision. pp. 640--658. Springer (2022)

\bibitem{cheng2021modular}
Cheng, H.K., Tai, Y.W., Tang, C.K.: Modular interactive video object segmentation: Interaction-to-mask, propagation and difference-aware fusion. In: Proceedings of the IEEE/CVF Conference on Computer Vision and Pattern Recognition. pp. 5559--5568 (2021)

\bibitem{cheng2021rethinking}
Cheng, H.K., Tai, Y.W., Tang, C.K.: Rethinking space-time networks with improved memory coverage for efficient video object segmentation. Advances in Neural Information Processing Systems  \textbf{34},  11781--11794 (2021)

\bibitem{cheng2018fast}
Cheng, J., Tsai, Y.H., Hung, W.C., Wang, S., Yang, M.H.: Fast and accurate online video object segmentation via tracking parts. In: Proceedings of the IEEE conference on computer vision and pattern recognition. pp. 7415--7424 (2018)

\bibitem{cheng2014global}
Cheng, M.M., Mitra, N.J., Huang, X., Torr, P.H., Hu, S.M.: Global contrast based salient region detection. IEEE transactions on pattern analysis and machine intelligence  \textbf{37}(3),  569--582 (2014)

\bibitem{cui2022mixformer}
Cui, Y., Jiang, C., Wang, L., Wu, G.: Mixformer: End-to-end tracking with iterative mixed attention. In: Proceedings of the IEEE/CVF Conference on Computer Vision and Pattern Recognition. pp. 13608--13618 (2022)

\bibitem{deng2009imagenet}
Deng, J., Dong, W., Socher, R., Li, L.J., Li, K., Fei-Fei, L.: Imagenet: A large-scale hierarchical image database. In: 2009 IEEE conference on computer vision and pattern recognition. pp. 248--255. Ieee (2009)

\bibitem{ding2023mose}
Ding, H., Liu, C., He, S., Jiang, X., Torr, P.H., Bai, S.: Mose: A new dataset for video object segmentation in complex scenes. arXiv preprint arXiv:2302.01872  (2023)

\bibitem{dosovitskiy2020image}
Dosovitskiy, A., Beyer, L., Kolesnikov, A., Weissenborn, D., Zhai, X., Unterthiner, T., Dehghani, M., Minderer, M., Heigold, G., Gelly, S., et~al.: An image is worth 16x16 words: Transformers for image recognition at scale. arXiv preprint arXiv:2010.11929  (2020)

\bibitem{duke2021sstvos}
Duke, B., Ahmed, A., Wolf, C., Aarabi, P., Taylor, G.W.: Sstvos: Sparse spatiotemporal transformers for video object segmentation. In: Proceedings of the IEEE/CVF Conference on Computer Vision and Pattern Recognition. pp. 5912--5921 (2021)

\bibitem{everingham2010pascal}
Everingham, M., Van~Gool, L., Williams, C.K., Winn, J., Zisserman, A.: The pascal visual object classes (voc) challenge. International journal of computer vision  \textbf{88},  303--338 (2010)

\bibitem{gao2022convmae}
Gao, P., Ma, T., Li, H., Lin, Z., Dai, J., Qiao, Y.: Convmae: Masked convolution meets masked autoencoders. arXiv preprint arXiv:2205.03892  (2022)

\bibitem{gao2022aiatrack}
Gao, S., Zhou, C., Ma, C., Wang, X., Yuan, J.: Aiatrack: Attention in attention for transformer visual tracking. In: European Conference on Computer Vision. pp. 146--164. Springer (2022)

\bibitem{guo2022adaptive}
Guo, P., Zhang, W., Li, X., Zhang, W.: Adaptive online mutual learning bi-decoders for video object segmentation. IEEE Transactions on Image Processing  \textbf{31},  7063--7077 (2022)

\bibitem{hariharan2011semantic}
Hariharan, B., Arbel{\'a}ez, P., Bourdev, L., Maji, S., Malik, J.: Semantic contours from inverse detectors. In: 2011 international conference on computer vision. pp. 991--998. IEEE (2011)

\bibitem{he2022masked}
He, K., Chen, X., Xie, S., Li, Y., Doll{\'a}r, P., Girshick, R.: Masked autoencoders are scalable vision learners. In: Proceedings of the IEEE/CVF conference on computer vision and pattern recognition. pp. 16000--16009 (2022)

\bibitem{hong2022lvos}
Hong, L., Chen, W., Liu, Z., Zhang, W., Guo, P., Chen, Z., Zhang, W.: Lvos: A benchmark for long-term video object segmentation. arXiv preprint arXiv:2211.10181  (2022)

\bibitem{hong2021adaptive}
Hong, L., Zhang, W., Chen, L., Zhang, W., Fan, J.: Adaptive selection of reference frames for video object segmentation. IEEE Transactions on Image Processing  \textbf{31},  1057--1071 (2021)

\bibitem{hu2018motion}
Hu, P., Wang, G., Kong, X., Kuen, J., Tan, Y.P.: Motion-guided cascaded refinement network for video object segmentation. In: Proceedings of the IEEE conference on computer vision and pattern recognition. pp. 1400--1409 (2018)

\bibitem{hu2018videomatch}
Hu, Y.T., Huang, J.B., Schwing, A.G.: Videomatch: Matching based video object segmentation. In: Proceedings of the European conference on computer vision (ECCV). pp. 54--70 (2018)

\bibitem{huang2020fast}
Huang, X., Xu, J., Tai, Y.W., Tang, C.K.: Fast video object segmentation with temporal aggregation network and dynamic template matching. In: Proceedings of the IEEE/CVF conference on computer vision and pattern recognition. pp. 8879--8889 (2020)

\bibitem{jang2016categorical}
Jang, E., Gu, S., Poole, B.: Categorical reparameterization with gumbel-softmax. arXiv preprint arXiv:1611.01144  (2016)

\bibitem{johnander2019generative}
Johnander, J., Danelljan, M., Brissman, E., Khan, F.S., Felsberg, M.: A generative appearance model for end-to-end video object segmentation. In: Proceedings of the IEEE/CVF conference on computer vision and pattern recognition. pp. 8953--8962 (2019)

\bibitem{khoreva2019lucid}
Khoreva, A., Benenson, R., Ilg, E., Brox, T., Schiele, B.: Lucid data dreaming for video object segmentation. International Journal of Computer Vision  \textbf{127}(9),  1175--1197 (2019)

\bibitem{li2022recurrent}
Li, M., Hu, L., Xiong, Z., Zhang, B., Pan, P., Liu, D.: Recurrent dynamic embedding for video object segmentation. In: Proceedings of the IEEE/CVF Conference on Computer Vision and Pattern Recognition. pp. 1332--1341 (2022)

\bibitem{li2018video}
Li, X., Loy, C.C.: Video object segmentation with joint re-identification and attention-aware mask propagation. In: Proceedings of the European conference on computer vision (ECCV). pp. 90--105 (2018)

\bibitem{lin2014microsoft}
Lin, T.Y., Maire, M., Belongie, S., Hays, J., Perona, P., Ramanan, D., Doll{\'a}r, P., Zitnick, C.L.: Microsoft coco: Common objects in context. In: Computer Vision--ECCV 2014: 13th European Conference, Zurich, Switzerland, September 6-12, 2014, Proceedings, Part V 13. pp. 740--755. Springer (2014)

\bibitem{lin2022swem}
Lin, Z., Yang, T., Li, M., Wang, Z., Yuan, C., Jiang, W., Liu, W.: Swem: Towards real-time video object segmentation with sequential weighted expectation-maximization. In: Proceedings of the IEEE/CVF Conference on Computer Vision and Pattern Recognition. pp. 1362--1372 (2022)

\bibitem{liu2021swin}
Liu, Z., Lin, Y., Cao, Y., Hu, H., Wei, Y., Zhang, Z., Lin, S., Guo, B.: Swin transformer: Hierarchical vision transformer using shifted windows. In: Proceedings of the IEEE/CVF international conference on computer vision. pp. 10012--10022 (2021)

\bibitem{maninis2018video}
Maninis, K.K., Caelles, S., Chen, Y., Pont-Tuset, J., Leal-Taix{\'e}, L., Cremers, D., Van~Gool, L.: Video object segmentation without temporal information. IEEE transactions on pattern analysis and machine intelligence  \textbf{41}(6),  1515--1530 (2018)

\bibitem{nowozin2014optimal}
Nowozin, S.: Optimal decisions from probabilistic models: the intersection-over-union case. In: Proceedings of the IEEE conference on computer vision and pattern recognition. pp. 548--555 (2014)

\bibitem{oh2018fast}
Oh, S.W., Lee, J.Y., Sunkavalli, K., Kim, S.J.: Fast video object segmentation by reference-guided mask propagation. In: Proceedings of the IEEE conference on computer vision and pattern recognition. pp. 7376--7385 (2018)

\bibitem{oh2019video}
Oh, S.W., Lee, J.Y., Xu, N., Kim, S.J.: Video object segmentation using space-time memory networks. In: Proceedings of the IEEE/CVF International Conference on Computer Vision. pp. 9226--9235 (2019)

\bibitem{perazzi2017learning}
Perazzi, F., Khoreva, A., Benenson, R., Schiele, B., Sorkine-Hornung, A.: Learning video object segmentation from static images. In: Proceedings of the IEEE conference on computer vision and pattern recognition. pp. 2663--2672 (2017)

\bibitem{perazzi2016benchmark}
Perazzi, F., Pont-Tuset, J., McWilliams, B., Van~Gool, L., Gross, M., Sorkine-Hornung, A.: A benchmark dataset and evaluation methodology for video object segmentation. In: Proceedings of the IEEE conference on computer vision and pattern recognition. pp. 724--732 (2016)

\bibitem{pont20172017}
Pont-Tuset, J., Perazzi, F., Caelles, S., Arbel{\'a}ez, P., Sorkine-Hornung, A., Van~Gool, L.: The 2017 davis challenge on video object segmentation. arXiv preprint arXiv:1704.00675  (2017)

\bibitem{rao2021dynamicvit}
Rao, Y., Zhao, W., Liu, B., Lu, J., Zhou, J., Hsieh, C.J.: Dynamicvit: Efficient vision transformers with dynamic token sparsification. Advances in neural information processing systems  \textbf{34},  13937--13949 (2021)

\bibitem{seong2020kernelized}
Seong, H., Hyun, J., Kim, E.: Kernelized memory network for video object segmentation. In: Computer Vision--ECCV 2020: 16th European Conference, Glasgow, UK, August 23--28, 2020, Proceedings, Part XXII 16. pp. 629--645. Springer (2020)

\bibitem{seong2021hierarchical}
Seong, H., Oh, S.W., Lee, J.Y., Lee, S., Lee, S., Kim, E.: Hierarchical memory matching network for video object segmentation. In: Proceedings of the IEEE/CVF International Conference on Computer Vision. pp. 12889--12898 (2021)

\bibitem{shi2015hierarchical}
Shi, J., Yan, Q., Xu, L., Jia, J.: Hierarchical image saliency detection on extended cssd. IEEE transactions on pattern analysis and machine intelligence  \textbf{38}(4),  717--729 (2015)

\bibitem{voigtlaender2019feelvos}
Voigtlaender, P., Chai, Y., Schroff, F., Adam, H., Leibe, B., Chen, L.C.: Feelvos: Fast end-to-end embedding learning for video object segmentation. In: Proceedings of the IEEE/CVF Conference on Computer Vision and Pattern Recognition. pp. 9481--9490 (2019)

\bibitem{voigtlaender2017online}
Voigtlaender, P., Leibe, B.: Online adaptation of convolutional neural networks for video object segmentation. arXiv preprint arXiv:1706.09364  (2017)

\bibitem{wang2021swiftnet}
Wang, H., Jiang, X., Ren, H., Hu, Y., Bai, S.: Swiftnet: Real-time video object segmentation. In: Proceedings of the IEEE/CVF Conference on Computer Vision and Pattern Recognition. pp. 1296--1305 (2021)

\bibitem{wang2023look}
Wang, J., Chen, D., Wu, Z., Luo, C., Tang, C., Dai, X., Zhao, Y., Xie, Y., Yuan, L., Jiang, Y.G.: Look before you match: Instance understanding matters in video object segmentation. In: Proceedings of the IEEE/CVF Conference on Computer Vision and Pattern Recognition. pp. 2268--2278 (2023)

\bibitem{wang2018semi}
Wang, W., Shen, J., Porikli, F., Yang, R.: Semi-supervised video object segmentation with super-trajectories. IEEE transactions on pattern analysis and machine intelligence  \textbf{41}(4),  985--998 (2018)

\bibitem{wu2023dropmae}
Wu, Q., Yang, T., Liu, Z., Wu, B., Shan, Y., Chan, A.B.: Dropmae: Masked autoencoders with spatial-attention dropout for tracking tasks. In: Proceedings of the IEEE/CVF Conference on Computer Vision and Pattern Recognition. pp. 14561--14571 (2023)

\bibitem{wu2023scalable}
Wu, Q., Yang, T., Wu, W., Chan, A.B.: Scalable video object segmentation with simplified framework. In: Proceedings of the IEEE/CVF International Conference on Computer Vision. pp. 13879--13889 (2023)

\bibitem{xiao2018monet}
Xiao, H., Feng, J., Lin, G., Liu, Y., Zhang, M.: Monet: Deep motion exploitation for video object segmentation. In: Proceedings of the IEEE Conference on Computer Vision and Pattern Recognition. pp. 1140--1148 (2018)

\bibitem{xu2018youtube}
Xu, N., Yang, L., Fan, Y., Yue, D., Liang, Y., Yang, J., Huang, T.: Youtube-vos: A large-scale video object segmentation benchmark. arXiv preprint arXiv:1809.03327  (2018)

\bibitem{xu2019mhp}
Xu, S., Liu, D., Bao, L., Liu, W., Zhou, P.: Mhp-vos: Multiple hypotheses propagation for video object segmentation. In: Proceedings of the IEEE/CVF Conference on Computer Vision and Pattern Recognition. pp. 314--323 (2019)

\bibitem{yang2022associating}
Yang, Z., Miao, J., Wang, X., Wei, Y., Yang, Y.: Associating objects with scalable transformers for video object segmentation. arXiv preprint arXiv:2203.11442  (2022)

\bibitem{yang2020collaborative}
Yang, Z., Wei, Y., Yang, Y.: Collaborative video object segmentation by foreground-background integration. In: Computer Vision--ECCV 2020: 16th European Conference, Glasgow, UK, August 23--28, 2020, Proceedings, Part V. pp. 332--348. Springer (2020)

\bibitem{yang2021collaborative}
Yang, Z., Wei, Y., Yang, Y.: Collaborative video object segmentation by multi-scale foreground-background integration. IEEE Transactions on Pattern Analysis and Machine Intelligence  \textbf{44}(9),  4701--4712 (2021)

\bibitem{yang2022decoupling}
Yang, Z., Yang, Y.: Decoupling features in hierarchical propagation for video object segmentation. arXiv preprint arXiv:2210.09782  (2022)

\bibitem{ye2022joint}
Ye, B., Chang, H., Ma, B., Shan, S., Chen, X.: Joint feature learning and relation modeling for tracking: A one-stream framework. In: European Conference on Computer Vision. pp. 341--357. Springer (2022)

\end{thebibliography}
\end{document}